\documentclass[10pt,twocolumn,letterpaper]{article}

\usepackage{cvpr}
\usepackage{times}
\usepackage{epsfig}
\usepackage{graphicx}
\usepackage{amsmath}
\usepackage{amssymb}
\usepackage{multirow}
\usepackage{booktabs}
\usepackage{array}
\usepackage{authblk}
\usepackage{footnote}
\usepackage{bm}
\makesavenoteenv{tabular}

\usepackage[breaklinks=true,bookmarks=false]{hyperref}

\cvprfinalcopy 

\def\cvprPaperID{****} 

\begin{document}

\title{The Solution for the CVPR 2023 1st foundation model challenge-Track2}
\author[1]{Haonan Xu}
\author[1]{Yurui Huang}
\author[1]{Sishun Pan}
\author[1]{Zhihao Guan}
\author[2]{Yi Xu}
\author[1]{{Yang Yang \thanks{Corresponding author: Yang Yang (yyang@njust.edu.cn)}}}
\affil[1]{Nanjing University of Science and Technology}
\affil[2]{Dalian University of Technology}

\maketitle
\begin{abstract}
In this paper, we propose a solution for cross-modal transportation retrieval. Due to the cross-domain problem of traffic images, we divide the problem into two sub-tasks of pedestrian retrieval and vehicle retrieval through a simple strategy. In pedestrian retrieval tasks, we use IRRA as the base model and specifically design an Attribute Classification to mine the knowledge implied by attribute labels. More importantly, We use the strategy of Inclusion Relation Matching to make the image-text pairs with inclusion relation have similar representation in the feature space. For the vehicle retrieval task, we use BLIP as the base model. Since aligning the color attributes of vehicles is challenging, we introduce attribute-based object detection techniques to add color patch blocks to vehicle images for color data augmentation. This serves as strong prior information, helping the model perform the image-text alignment. At the same time, we incorporate labeled attributes into the image-text alignment loss to learn fine-grained alignment and prevent similar images and texts from being incorrectly separated. Our approach ranked first in the final B-board test with a score of 70.9.
\end{abstract}

\section{Introduction}

With the development of multimodal large-scale model technologies \cite{yang2019semi, yang2019semi1, yang2018complex, yang2019deep,yang2021corporate}, the unification of text and image representations and modality conversion has been widely applied \cite{fu2024noise,wancovlr,yang2021corporate,yang2021learning,wancovlr,yang2021cost,yang2021s2osc}, which can further improve the accuracy and flexibility of image retrieval. In this competition, we found through experiments that there is a difficulty in cross-domain learning when unifying pedestrian and vehicle modeling. Therefore, we designed simple rule matching and classifiers to divide the task into two sub-tasks: pedestrian retrieval and vehicle retrieval. To accomplish the two sub-tasks, we use IRRA\cite{jiang2023cross} as the base model for pedestrian retrieval. This model adopts the CLIP\cite{radford2021learning} base architecture and designs an implicit relationship inference module to learn the relationships between local visual-text annotations and enhance global image-text matching. We use BLIP\cite{li2022blip} as the base model for vehicle retrieval, which is pre-trained on 120 million image-text pairs and has strong image-text alignment capabilities.

\begin{figure}
\begin{center}
\includegraphics[width=0.5\textwidth]{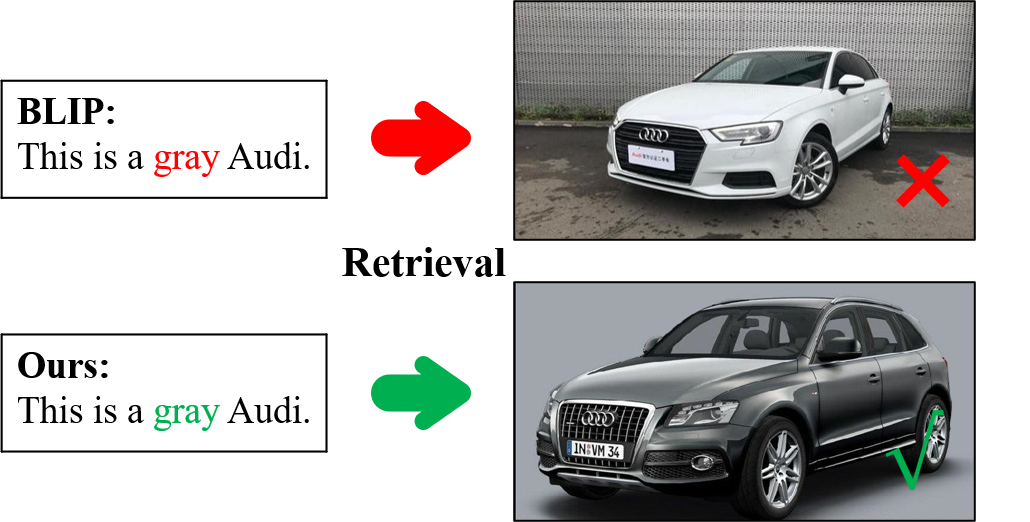}
\end{center}
   \caption{This example illustrates the improvement of using a visual prompt augmentation strategy.}
\label{fig:short}
\end{figure}

\begin{figure}
\begin{center}
\includegraphics[width=0.5\textwidth]{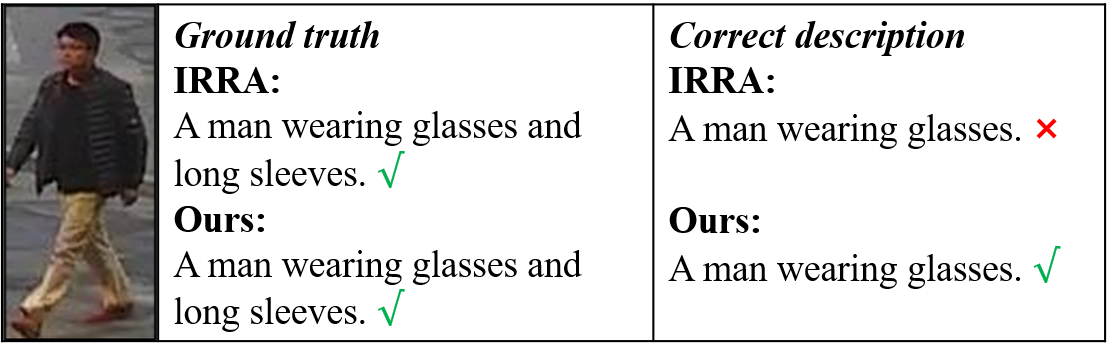}
\end{center}
   \caption{The ground truth text of the image is described as "A man wearing glasses and long sleeves.", for which both our model and IRRA are correctly retrieved. However, when the text is "A man wearing glasses.", our model can retrieve the image, while IRRA cannot.}
\label{fig:short}
\end{figure}

The main contributions can be summarized as follows:

$\bullet$ \textbf{Multi-label classification of pedestrian attributes.} For the pedestrian retrieval task, we use Attribute Classification to fully mine pedestrian category information. 

$\bullet$ \textbf{Vehicle image color augmentation based on object detection.} We provide the model with prior knowledge of colors to help it be more accurate in color recognition. Refer to Figure 1 for a specific example explanation.

$\bullet$ \textbf{Fine-grained alignment.} Considering that in the process of contrastive learning, there often exist multiple image-text pairs of the same category in the same batch, our proposed strategy makes the image-text pairs with the same category attribute label(inclusion relation in pedestrian retrieval tasks, refer to Figure 2 ) in the same batch have similar feature representation. This strategy significantly improves retrieval accuracy.


\begin{figure*}
\begin{center}
\includegraphics[width=1.05\textwidth]{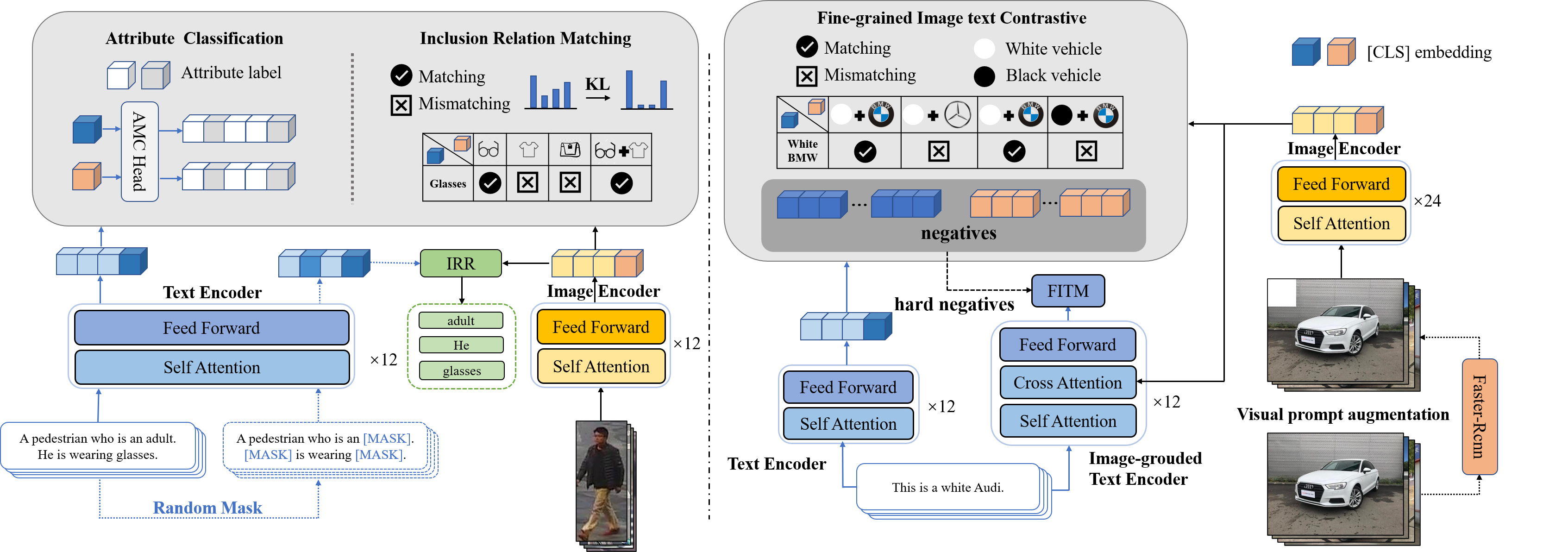}
\end{center}
   \caption{Overall Architecture. On the left of the figure is the pedestrian model architecture. On the basis of IRRA, Attribute Classification, and Inclusion Relation Matching are innovatively added. The design of the model fully mines the data annotation information and makes the performance of image-text pairs more excellent in fine-grained matching. The right side of the figure shows the vehicle model architecture. We adopt BLIP as a framework to optimize the fine-grained matching of image-text pairs and mine difficult samples. We also introduce a visual prompt augmentation strategy to give the model prior knowledge to help the model better recognize the vehicle color. }
\label{fig:short}
\end{figure*}

\section{Related Work}

\textbf{Vision-Language Pre-training} aims to learn the semantic correspondence between visual patterns and language patterns by pre-training on large-scale image-text pairs. Inspired by the success of Transformer-based\cite{vaswani2017attention} language model pre-training (such as BERT)\cite{devlin2018bert}, and Vision Transformer (ViT)\cite{dosovitskiy2020image}. Vision-Language Pre-training (VLP) has emerged as the prevailing paradigm in learning multimodal representations. And large-scale vision-language Transformers have stimulated deeper modality interactions in retrieval models.

\textbf{Text-to-image retrieval of pedestrian and vehicle} is a novel retrieval task aimed at inputting open text to retrieve relevant pedestrian or vehicle images. \cite{jiang2023cross} implements pedestrian retrieval from text to image by designing an implicit relationship inference module, which adopts the masked language modeling paradigm. This module can learn relationships between local visual-text annotations and enhance global image-text matching without requiring additional prior supervision. In previous work, vehicle retrieval from text to image has not been achieved\cite{he2020multi, liu2017provid}. Past research focused on vehicle re-identification tasks, i.e., image-to-image retrieval. Since pedestrians and vehicles belong to two different domains, it is necessary to address the problem of cross-domain retrieval.

\section{Method}
\subsection{Overall Architecture}

Figure 3 illustrates the overall framework of our solution. During the training stage, we train the pedestrian and vehicle models separately. In the testing stage, we obtain the final results by merging the results from both pedestrian and vehicle models. Next, we will introduce the details of the pedestrian module and the vehicle module specifically.

\subsection{Pedestrian Retrieval}

In this section, we introduce the pedestrian text-to-image retrieval scheme. We adopt the IRRA framework and make adaptive improvements. Next, we will specifically introduce the scheme.

\textbf{Cross-Modal Implicit Relation Reasoning (IRR).} We perform modality fusion on masked text features and image features to predict the words that are masked in the text. This approach achieves a fine-grained association between text and images.

\textbf{Attribute Classification (AC).} Pedestrian attribute labels are used for multi-label classification of text and image features. It explicitly takes intra-modal distances into account and ensures that feature representations of images/texts with the same attribute are closely clustered together in the joint embedding space. We select [CLS] tokens as representations for image and text, and then a fully connected layer with shared weights maps them to q classes for multi-label classification.
The objective of AC can be described as follows:
\begin{equation}
\mathcal{L}_{AC} = \frac{1}{2}\mathbb{E}_{(i,t)}(CE(A,\hat{A}_{i})+CE(A,\hat{A}_{t}))
\end{equation}
where $ \hat{A}_{i} $ and $ \hat{A}_{t} $ denote the prediction attribute for text and image, respectively. $  A $ is the ground-truth attribute label. $ CE $ denotes cross entropy loss.

We propose a novel cross-modal Matching loss, called \textbf{Inclusion Relation Matching (IRM)}. Let sim(u, v) denote the dot product between $ L_2 $ normalized u and v(i.e. cosine similarity). The cosine similarity distributions of images and texts are embedded in the KL divergence to correlate the representations of different modalities.
The image-text matching probability is computed by:
\begin{equation}
p_{i,j}=\frac{\exp{(sim(f^{txt}_i,f^{img}_j)/\tau})}{\sum_{k=1}^{B}\exp({sim(f^{txt}_i,f^{img}_k)/\tau)}}    
\end{equation}
where $ f^{txt} $ and $ f^{img} $ denote the [CLS] token for text and image, respectively. $ \tau $ is the temperature scale parameter. 

Due to the particularity of the pedestrian retrieval task, we need to retrieve images that contain textual information. Therefore, Image-text pairs with inclusion relations should have similar feature representations in the feature space. In order to increase the matching probability of image-text pairs with inclusion relations, our proposed matching probability can be expressed as follows:

\begin{equation}
\begin{split}
q_{i,j}=\frac{a_{i,j}}{\sum_{k=1}^{B}a_{i,k}}, \quad a_{i,j}=\begin{cases}
    1& \text{ if } A_i\subseteq A_j\\
    0&  otherwise
\end{cases}
\end{split}
\end{equation}

Then IRM loss in a mini-batch can be calculated as follows:
\begin{equation}
\begin{split}
\mathcal{L}_{IRM} = \mathbb{E}_{i}(KL(p^{i2t}_i||q_i)+KL(p^{t2i}_i||q_i))\\
KL(p_i||q_i)=\frac{1}{B}\sum_{j=1}^{B}p_{i,j}\log{(\frac{p_{i,j}}{q_{i,j}+\epsilon})}
\end{split}
\end{equation}
where $ \epsilon $ is a small number to avoid numerical problems, and $ B $ is the size of a mini-batch.

The final pedestrian retrieval task scheme is optimized with the following objectives:

\begin{equation}
\mathcal{L}_{p} = \mathcal{L}_{IRR}+\mathcal{L}_{AC}+\mathcal{L}_{IRM}    
\end{equation}

\subsection{Vehicle Retrieval}

In this section, the scheme of vehicle retrieval will be introduced, the vehicle retrieval task adopts the BLIP architecture and uses BLIP-Large as the initialization of the network. To better optimize the vehicle retrieval task, we propose a color augmentation of vehicle images based on attribute object detection and fine-grained image-text contrastive learning.

\textbf{Color Prompt Augmentation of Vehicle Image.} In our experiments, we found that without image augmentation processing, the model is easy to confuse some colors, such as white and gray, which leads to incorrect retrieval results. We use the object detection method with attribute detection \cite{han2021image} to detect the color attribute of the vehicle. In order to better help the model understand the color attribute, we add the predicted RGB color patch on the top left corner of the vehicle image as prior information to help the model better understand the color information of the vehicle. The whole process is shown in Figure 3. Experiments show that the augmentation method can correct the wrong cases of color retrieval and greatly improve overall performance.

\textbf{Fine-grained Image-Text Contrastive Loss (FITC).} The traditional ITC loss pulls the correct image-text pair closer and pushes the wrong image-text pair far away. In this task, the image-text pair has a many-to-many relationship. Due to the image-text pair having label attributes, such as white Audi, so image-text pairs of the same type should have similar feature representations. Therefore, the category label is introduced into contrastive learning to make the same kind of image-text as close as possible in the semantic space. For each image and text, we calculate the softmax-normalized image-to-text and text-to-image similarity as:

\begin{equation}
\begin{split}
p^{i2t}_{m}\left (I\right)=\frac{\exp\left(s\left (I,T_{m}\right){/}\tau\right)}{{\textstyle \sum_{m=1}^{M}} \exp\left(s\left(I,T_{m}\right ){/}\tau\right)},\\
p^{t2i}_{m}\left (T\right)=\frac{\exp\left(s\left(T, I_{m}\right ){/}\tau\right)}{{\textstyle\sum_{m=1}^{M}}\exp\left(s\left(T, I_{m}\right){/}\tau\right)} 
\end{split}
\end{equation}
where $\tau$ is a learnable temperature parameter, and $s\left ( \cdot ,\cdot  \right ) $ is the cos function.  Let $\bm{y} ^{i2t}\left ( I \right ) $ and $\bm{y} ^{t2i}\left ( T \right ) $ denote the ground-truth similarity. And $\bm{y} ^{i2t}\left ( I \right ) = \bm{y} ^{t2i}\left ( T \right ) = \left[\frac{1}{K}, \frac{1}{K}, ...,0\right]$, and $K$ is the number of same tags. The image-text contrastive loss is defined as the cross-entropy H between $\bm{p}$ and $\bm{y}$:


\begin{equation}
\begin{split}
\mathcal{L}_{FITC}=\frac{1}{2}\mathbb{E}_{\left(I,T\right)\sim D}\left[H\left(\bm{y}^{i2t}\left(I\right),\bm{p}^{i2t}\left(I\right)\right) + H\left(\bm{y}^{t2i}\left(T\right),\bm{p}^{t2i}\left(T\right)\right)\right]
\end{split}
\end{equation}

\textbf{Fine-grained Image-Text Matching (FITM) Loss.} Traditional ITM is a binary classification task, where the model uses ITM heads (linear layers) to predict whether a multimodal feature for a given image-text pair is positive (matched) or negative (mismatched). To find more informative negative pairs, we adopt the hard negative pair mining strategy of \cite{li2021align}. Negative pairs with higher contrastive similarity are more likely to be selected to compute the loss. Similarly, we introduce category labels in negative example mining, so that similar image-text pairs will not be negative pairs. The FITM loss is:

\begin{equation}
\mathcal{L}_{FITM} = \mathbb{E}_{\left ( I,T \right )\sim D} H\left ( \bm{y} ^{itm}, \bm{p} ^{itm} \left ( I,T \right ) \right )
\end{equation}
where $\bm{y}_{itm}$ is a 2-dimensional one-hot vector representing the ground-truth label.

The final vehicle retrieval training objective is:
\begin{equation}
\mathcal{L}_{v} = \mathcal{L}_{FITC}+\mathcal{L}_{FITM}    
\end{equation}

\section{Experiments}
\textbf{Dataset.} The dataset was provided by the competition officials. The data are presented in the form of image-text pairs. The training data includes 90000 pedestrian samples and 46,117 vehicle samples. The test data contains 10000 samples for pedestrian data and 7611 samples for vehicle data.

\textbf{Implementation Detail.} The model for the pedestrian retrieval task consists of a pre-trained image encoder, a pre-trained text encoder, and a random-initialized multimodal interaction encoder. In terms of vehicle retrieval tasks, We use BLIP and initialized it with BLIP-Large, which is pre-trained on 120 million image-text pairs. We only use a single NVIDIA RTX 4090 to train the pedestrian model, we set the initial learning rate as 1e-5, the batch size as 90, and the epoch as 60. We train the vehicle retrieval model using NVIDIA RTX 4090×4 with an initial learning rate of 3e-5, the batch size is set to 32 and the epoch is set to 6.

\textbf{Result.} Table 1 shows the performance of applying our method on the test set. We add our methods step by step in the order of the labels in the table. The experimental results well confirm the effectiveness of our proposed method. Mining attribute information and visual prompt augmentation strategies effectively help the model learn knowledge better. More importantly, the effectiveness of the fine-grained matching method significantly improves the retrieval accuracy of the model.

\begin{table}[!htbp]
\caption{Results of our method on the test set.}
\label{table:1}
\begin{center}
\begin{tabular}{ >{\centering\arraybackslash}m{0.05cm} 
>{\centering\arraybackslash}m{6.1cm} 
>{\centering\arraybackslash}m{0.7cm} 
}
 \hline
 \# & Method &  Score \\ 
 \hline
 1 & IRRA + BLIP & $ 0.72+ $\\ 
 2 & Inclusion Relation Matching & $ 0.76+ $\\ 
 3 & Attribute classification, color augmentation  & $ 0.79+ $\\ 
 4 & Fine-grained strategies for vehicle & $ 0.82+$ \\
 \textbf{5} & \textbf{Final result (B board)} & $\textbf{0.709+}$ \\ 
\hline
\end{tabular}
\end{center}
\end{table}

\section{Conclusion}
This report summarizes our solution for the CVPR 2023 1st foundation model challenge-Track2. Our solution indicates that incorporating attribute information into the contrastive learning process can help the image text features align better, we introduce visual prompt augmentation and Cross-Modal Implicit Relation Reasoning methods to help the model better perceive the fine-grained relation. The final competition results show the effectiveness of our solution.

{\small
\bibliographystyle{ieee}
\bibliography{egpaper_final}

\begin{thebibliography}{10}\itemsep=-1pt

\bibitem{devlin2018bert}
J.~Devlin, M.-W. Chang, K.~Lee, and K.~Toutanova.
\newblock Bert: Pre-training of deep bidirectional transformers for language
  understanding.
\newblock {\em arXiv preprint arXiv:1810.04805}, 2018.

\bibitem{dosovitskiy2020image}
A.~Dosovitskiy, L.~Beyer, A.~Kolesnikov, D.~Weissenborn, X.~Zhai,
  T.~Unterthiner, M.~Dehghani, M.~Minderer, G.~Heigold, S.~Gelly, et~al.
\newblock An image is worth 16x16 words: Transformers for image recognition at
  scale.
\newblock {\em arXiv preprint arXiv:2010.11929}, 2020.

\bibitem{fu2024noise}
Z.~Fu, K.~Song, L.~Zhou, and Y.~Yang.
\newblock Noise-aware image captioning with progressively exploring mismatched
  words.
\newblock In {\em Proceedings of the AAAI Conference on Artificial
  Intelligence}, volume~38, pages 12091--12099, 2024.

\bibitem{han2021image}
X.~Han, J.~Yang, H.~Hu, L.~Zhang, J.~Gao, and P.~Zhang.
\newblock Image scene graph generation (sgg) benchmark.
\newblock {\em arXiv preprint arXiv:2107.12604}, 2021.

\bibitem{he2020multi}
S.~He, H.~Luo, W.~Chen, M.~Zhang, Y.~Zhang, F.~Wang, H.~Li, and W.~Jiang.
\newblock Multi-domain learning and identity mining for vehicle
  re-identification.
\newblock In {\em Proceedings of the IEEE/CVF Conference on Computer Vision and
  Pattern Recognition Workshops}, pages 582--583, 2020.

\bibitem{jiang2023cross}
D.~Jiang and M.~Ye.
\newblock Cross-modal implicit relation reasoning and aligning for
  text-to-image person retrieval.
\newblock In {\em Proceedings of the IEEE/CVF Conference on Computer Vision and
  Pattern Recognition}, pages 2787--2797, 2023.

\bibitem{li2022blip}
J.~Li, D.~Li, C.~Xiong, and S.~Hoi.
\newblock Blip: Bootstrapping language-image pre-training for unified
  vision-language understanding and generation.
\newblock In {\em International Conference on Machine Learning}, pages
  12888--12900. PMLR, 2022.

\bibitem{li2021align}
J.~Li, R.~Selvaraju, A.~Gotmare, S.~Joty, C.~Xiong, and S.~C.~H. Hoi.
\newblock Align before fuse: Vision and language representation learning with
  momentum distillation.
\newblock {\em Advances in neural information processing systems},
  34:9694--9705, 2021.

\bibitem{liu2017provid}
X.~Liu, W.~Liu, T.~Mei, and H.~Ma.
\newblock Provid: Progressive and multimodal vehicle reidentification for
  large-scale urban surveillance.
\newblock {\em IEEE Transactions on Multimedia}, 20(3):645--658, 2017.

\bibitem{radford2021learning}
A.~Radford, J.~W. Kim, C.~Hallacy, A.~Ramesh, G.~Goh, S.~Agarwal, G.~Sastry,
  A.~Askell, P.~Mishkin, J.~Clark, et~al.
\newblock Learning transferable visual models from natural language
  supervision.
\newblock In {\em International conference on machine learning}, pages
  8748--8763. PMLR, 2021.

\bibitem{vaswani2017attention}
A.~Vaswani, N.~Shazeer, N.~Parmar, J.~Uszkoreit, L.~Jones, A.~N. Gomez,
  {\L}.~Kaiser, and I.~Polosukhin.
\newblock Attention is all you need.
\newblock {\em Advances in neural information processing systems}, 30, 2017.

\bibitem{wancovlr}
F.~Wan, W.~Xiangyu, Z.~Guan, and Y.~Yang.
\newblock Covlr: Coordinating cross-modal consistency and intra-modal relations
  for vision-language retrieval.

\bibitem{yang2019semi1}
Y.~Yang, Z.-Y. Fu, D.-C. Zhan, Z.-B. Liu, and Y.~Jiang.
\newblock Semi-supervised multi-modal multi-instance multi-label deep network
  with optimal transport.
\newblock {\em IEEE Transactions on Knowledge and Data Engineering},
  33(2):696--709, 2019.

\bibitem{yang2021learning}
Y.~Yang, Z.-Q. Sun, H.~Zhu, Y.~Fu, Y.~Zhou, H.~Xiong, and J.~Yang.
\newblock Learning adaptive embedding considering incremental class.
\newblock {\em IEEE Transactions on Knowledge and Data Engineering},
  35(3):2736--2749, 2021.

\bibitem{yang2021s2osc}
Y.~Yang, H.~Wei, Z.-Q. Sun, G.-Y. Li, Y.~Zhou, H.~Xiong, and J.~Yang.
\newblock S2osc: A holistic semi-supervised approach for open set
  classification.
\newblock {\em ACM Transactions on Knowledge Discovery from Data (TKDD)},
  16(2):1--27, 2021.

\bibitem{yang2018complex}
Y.~Yang, Y.-F. Wu, D.-C. Zhan, Z.-B. Liu, and Y.~Jiang.
\newblock Complex object classification: A multi-modal multi-instance
  multi-label deep network with optimal transport.
\newblock In {\em Proceedings of the 24th ACM SIGKDD International Conference
  on Knowledge Discovery \& Data Mining}, pages 2594--2603, 2018.

\bibitem{yang2019deep}
Y.~Yang, Y.-F. Wu, D.-C. Zhan, Z.-B. Liu, and Y.~Jiang.
\newblock Deep robust unsupervised multi-modal network.
\newblock In {\em Proceedings of the AAAI Conference on Artificial
  Intelligence}, volume~33, pages 5652--5659, 2019.

\bibitem{yang2021corporate}
Y.~Yang, J.-Q. Yang, R.~Bao, D.-C. Zhan, H.~Zhu, X.-R. Gao, H.~Xiong, and
  J.~Yang.
\newblock Corporate relative valuation using heterogeneous multi-modal graph
  neural network.
\newblock {\em IEEE Transactions on Knowledge and Data Engineering},
  35(1):211--224, 2021.

\bibitem{yang2019semi}
Y.~Yang, D.-C. Zhan, Y.-F. Wu, Z.-B. Liu, H.~Xiong, and Y.~Jiang.
\newblock Semi-supervised multi-modal clustering and classification with
  incomplete modalities.
\newblock {\em IEEE Transactions on Knowledge and Data Engineering},
  33(2):682--695, 2019.

\bibitem{yang2021cost}
Y.~Yang, D.-W. Zhou, D.-C. Zhan, H.~Xiong, Y.~Jiang, and J.~Yang.
\newblock Cost-effective incremental deep model: Matching model capacity with
  the least sampling.
\newblock {\em IEEE Transactions on Knowledge and Data Engineering},
  35(4):3575--3588, 2021.

\end{thebibliography}
}

\end{document}